\documentclass[letterpaper]{article} 
\usepackage{aaai23}
\usepackage{times}  
\usepackage{helvet}  
\usepackage{courier}  
\usepackage[hyphens]{url}  
\usepackage{graphicx} 
\newcommand\keywords[1]{\textbf{Keywords}: #1}
\usepackage[square,numbers]{natbib}  
\usepackage{caption} 
\usepackage{algorithm}
\usepackage{algorithmic}

\usepackage{newfloat}
\usepackage{listings}
\DeclareCaptionStyle{ruled}{labelfont=normalfont,labelsep=colon,strut=off} 
\floatstyle{ruled}
\newfloat{listing}{tb}{lst}{}
\floatname{listing}{Listing}
\title{TSAA: A Two-Stage Anchor Assignment Method towards Anchor Drift in Crowded Object Detection}

\author {
    Li Xiang\textsuperscript{\rm 1,2,3,4,*},
    He Miao \textsuperscript{\rm 1,2,3},
    Luo Haibo \textsuperscript{\rm 1,2,3},
    Yang Huiyuan \textsuperscript{\rm 5},
    Xiao Jiajie \textsuperscript{\rm 6},
}

\affiliations {
    \textsuperscript{\rm 1} Key Laboratory of Opto-Electronic Information Processing, Chinese Academy of Sciences\\
    \textsuperscript{\rm 2} Shenyang Institute of Automation, Chinese Academy of Sciences\\
    \textsuperscript{\rm 3} Institutes for Robotics and Intelligent Manufacturing, Chinese Academy of Sciences\\
    \textsuperscript{\rm 4} University of Chinese Academy of Sciences\\
    \textsuperscript{\rm 5} McGill University, Montreal,Canada\\
    \textsuperscript{\rm 6} Trip.com Group\\
     \textsuperscript{\rm *} Corresponding Author. Email: lixiang184@mails.ucas.ac.cn.
}

\usepackage{bibentry}

\begin{document}

\maketitle

\begin{abstract}
Among current anchor-based detectors, a positive anchor box will be intuitively assigned to the object that overlaps it the most. The assigned label to each anchor will directly determine the optimization direction of the corresponding prediction box, including the direction of box regression and category prediction. In our practice of crowded object detection, however, the results show that a positive anchor does not always regress toward the object that overlaps it the most when multiple objects overlap. We name it \textit{anchor drift}. The \textit{anchor drift} reflects that the anchor-object matching relation, which is determined by the degree of overlap between anchors and objects, is not always optimal. Conflicts between the fixed matching relation and learned experience in the past training process may cause ambiguous predictions and thus raise the false-positive rate. In this paper, a simple but efficient adaptive two-stage anchor assignment (TSAA) method is proposed. It utilizes the final prediction boxes rather than the fixed anchors to calculate the overlap degree with objects to determine which object to regress for each anchor. The participation of the prediction box makes the anchor-object assignment mechanism adaptive. Extensive experiments are conducted on three classic detectors RetinaNet, Faster-RCNN and YOLOv3 on CrowdHuman and COCO to evaluate the effectiveness of TSAA. The results show that TSAA can significantly improve the detectors’ performance without additional computational costs or network structure changes.
\end{abstract}

\keywords{Crowded Object Detection, Pedestrian Detection, Anchor-based Detector, Convolutional Neural Networks, Deep Learning.}

\section{Introduction} 

Object detection is an essential task in computer vision. As one of the most common methods, anchor-based models have achieved state-of-the-art (SOTA) performance in many branch research fields such as crowded object detection\cite{bib1,bib2,bib3}, remote sensing target detection\cite{bib4,bib5}, small target detection\cite{bib6,bib7, bib47, bib48}, pedestrian detection\cite{bib8,bib9,bib10,bib11,bib12, bib46}, etc. In anchor-based networks, dense anchor boxes paved across the feature maps ensure that the objects with random locations, shapes, and scales can be captured by the detector as much as possible. 
\begin{figure}[t]
\centering
\includegraphics[width=0.9\columnwidth]{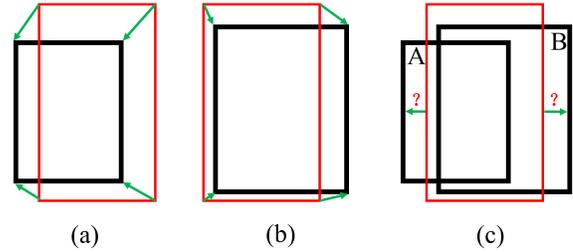} 
\caption{Difference of regression directions between sparse and dense cases. (a)(b) There is only one regression option when the target is sparse and has no contact with others. (c) The choice of regression direction becomes ambiguous when targets overlap}
\label{fig1}
\end{figure}
In classic anchor-based methods, such as Faster RCNN\cite{bib13}, YOLO series\cite{bib14,bib15,bib16,bib17,bib18}, SSD\cite{bib19}, and RetinaNet\cite{bib20}, anchor boxes are firstly judged as positive or negative samples, then the positive ones are uniquely assigned to one of the objects. This matching relationship is defined according to Intersection over Union (IoU) between the preset anchor boxes and object boxes. Once one anchor box matches one unique object, this matching relationship will be kept during optimization.However, whether this fixed but stiff matching relation is optimal for anchor-based detectors has hardly been discussed in previous works. 

According to the observation, in crowded scenes, some anchor boxes do not always regress to the objects they are initially assigned to, which is called \textit{anchor drift}. The \textit{anchor drift} can confuse the prediction head and will make the prediction boxes incompact. When the objects are sparse, i.e., objects do not overlap others, the prediction head will learn a kind of regression pattern, as shown in Fig1(a) and Fig1(b), the red anchor box will respectively regress to the objects as the green arrow indicated, because there is only one choice for the anchor to regress. However, the situation will be changed when objects become crowded. As shown in Fig1(c), although the anchor box will still be judged as positive when objects A and B overlap, the prediction head will be confused about which object to regress for each anchor? For example, when the anchor box is assigned to object A but the prediction head decides to follow the regression pattern learned in Fig1(b), the anchor will drift since the preset optimization target is different from the actual situation. Thus the prediction head will be confused because the optimization forces the anchor to regress to object A, which will conflict with the pattern learned from the situation in Fig1(b). This conflict probably causes the final prediction boxes to neither regress to object A nor to object B precisely and will generate an ambiguous prediction box which will increase the incompactness of prediction boxes. 

In this paper, we propose a two-stage anchor assignment (TSAA) method to untie the binding relationship between anchors and objects during the optimization process and give the positive anchors another chance to rematch one of the objects adaptively. It is well known that the predicted coordinate offsets and the anchor boxes are in one-to-one correspondence. Before the prediction output, we integrate the predicted coordinate offsets with the corresponding anchor boxes and restore the prediction boxes into the image coordinate system. Then we calculate the IoU between the prediction boxes and the object boxes and assign the prediction boxes to the objects with the maximum IoU respectively, which means each anchor box is simultaneously assigned to the object it is most likely to match since the anchor boxes and prediction boxes are one-to-one. According to the new matching relationship, we calculate the box coordinate offset targets and define category labels to optimize the networks. The participation of the prediction box makes the anchor-object assignment mechanism adaptive. This adaptive anchor-object reassignment can avoid the conflict shown in Fig1(c) and reduce the ambiguous predictions. 

Extensive experiments are conducted on classic anchor-based detectors. The experimental results show that the TSAA method can decrease the Missing Rate on RetinaNet, Faster RCNN with FPN, and YOLOv3 by 4.35\%, 0.81\%, and 0.19\% on CrowdHuman\cite{bib44}, respectively, which means the TSAA can effectively suppress the false-positive predictions. The contributions can be summarized as follows:
\begin{itemize}
	\item The \textit{anchor drift} phenomenon, which can increase the false-positive rate of anchor-based detectors in crowded scenes, is discovered and introduced.
	\item A two-stage anchor assignment method is proposed to solve the anchor drift problem.
	\item Extensive experiments are conducted to demonstrate that the TSAA method can effectively solve the anchor drift phenomenon and decrease ambiguous predictions to enhance the compactness of prediction boxes.

\end{itemize}

\section{Related Work}
\subsection{Generic Object Detection}
With the development of deep learning, convolutional neural network(CNN) based detectors have become SOTA in computer vision. RCNN\cite{bib21} was the first detector to apply the CNN, based on which the Fast RCNN\cite{bib22} was proposed to extract information for each proposal box from the deep feature. Then, Faster RCNN presented Region Proposal Network (RPN) to generate high-quality proposals, and Region of Interest (RoI) pooling was utilized to generate fixed-size output. Faster RCNN established the general framework of two-stage detectors. Subsequently, the PSRoI pooling\cite{bib25}, RoIAlign\cite{bib23}, PrRoI pooling\cite{bib24} were proposed to improve RoI pooling process. FPN\cite{bib26} was proposed to fuse deep and shallow features to improve the object scale variety problem. Similar to the RPN, one-stage detectors pave dense anchor boxes across feature maps and directly output dense predictions based on each anchor box. SSD proposed a one-stage framework that utilizes multi-scale feature maps to detect multi-scale objects. RetinaNet proposed a focal loss to make the training process pay more attention to the hard samples. YOLO series utilized the K-means algorithm to count the scales and ratios distribution of target boxes in datasets and found 9 clusters as anchors. \cite{bib29} proposed an adaptive anchor boxes optimization method AABO, which was based on Bayesian optimization and the Sub-Sample method and could automatically and efficiently search for optimal anchor settings. 

On account of anchor settings will dramatically influence the performance of detectors, anchor-free detectors have been proposed recently to avoid the negative impact brought by anchors. CornerNet\cite{bib30} and CenterNet\cite{bib31} were keypoint-based methods to represent target boxes by corner points and center points, respectively. FCOS\cite{bib32} utilized multi-scale features to solve the ambiguity when objects overlap each other. 

\subsection{Crowded Object Detection}
In general, two reasons make it difficult to detect crowded objects compared with sparse scenes: 1) similar features of highly overlapped instances are indistinguishable from each other; 2) heavily overlapped prediction boxes are easily suppressed during the NMS process. \cite{bib33} proposed a novel concept that each proposal predicts multiple rather one objects to solve the problem of feature confusion problem. Meanwhile, a set-NMS was proposed to preserve the prediction boxes generated based on the same proposals during the traditional NMS process. \cite{bib34} followed the iterative scheme to detect a subset of objects at each iteration, and there were no interactions between iterations. \cite{bib35} introduced a crowded object feature extraction module, which fuses the spatial pyramid and the pixel shuffle module to boost YOLO detectors’ distinguishable feature extraction ability. \cite{bib36,bib37,bib38,bib39} paid attention to the improvements in the NMS process. Soft NMS decreased the confidence score according to overlap degree rather than suppressing heavily overlapped boxes directly as the hard NMS. Adaptive NMS predicted the density of each object and suppresses all boxes whose overlap degree are greater than the density. R$^2$NMS predictd the visible boxes and uses that to guide the NMS process since the visible boxes rarely overlap each other. NOH-NMS predictd the relative position information of another object box with the largest overlap based on each prediction box and used the relative information to guide the NMS process. In addition, the bad compactness of prediction boxes will increase the false-positive rate in crowded scenes. \cite{bib40} proposed an aggregation loss to enforce the proposals located closely and compactly to the ground-truth object. \cite{bib41} introduced a novel box regression loss named repulsion loss, whose motivation was to make the prediction box attracted by its target and repulsed by other surrounding objects.

\section{Analysis of Anchor Drift}
In current anchor-based detectors, anchor boxes are manually defined and fixed. Each anchor box will be judged as positive or negative, and the positive ones will be uniquely assigned to one of the objects. Once the positive anchor-object matching relationship is defined, the optimizer will force the network to satisfy that relationship. This matching relationship is significant because it determines which object each output corresponds to and which object will dominate the feature extraction. In general object detection, i.e., there is rare overlap between objects, it is reasonable that each anchor box should be assigned to the object with the maximum IoU value. However, when the objects become crowded, the situation will be changed.

In the experiment on crowded object detection, we output the IoU values between the prediction boxes that correspond to the same objects during the training process of an anchor-based detector. In expectation,these IoU values should all tend to 1 since their defined regression destinations are the same. However, the actual results are different. Although some anchors are assigned to the same objects, the IoU values between their corresponding prediction boxes are possibly closer to 0 than 1, which means some of these anchors do not regress to the objects they were initially assigned to.AS shown in Fig2, we count the minimum IoU of target-prediction-pair (MITP) in each image during training process and output mean MITP of each epoch. The most loose prediction box in an image can reflect the lower bound of the prediction boxes’ compactness. In Fig2 we can see that the most loose prediction box generated by the original RetinaNet may be very far away from its target box. Since the numbers in Fig2 are mean MITP in each epoch, the MITP in certain image can be smaller than the mean values , even tends to 0.
 
\begin{figure}[thp]
\centering
\includegraphics[width=0.8\columnwidth]{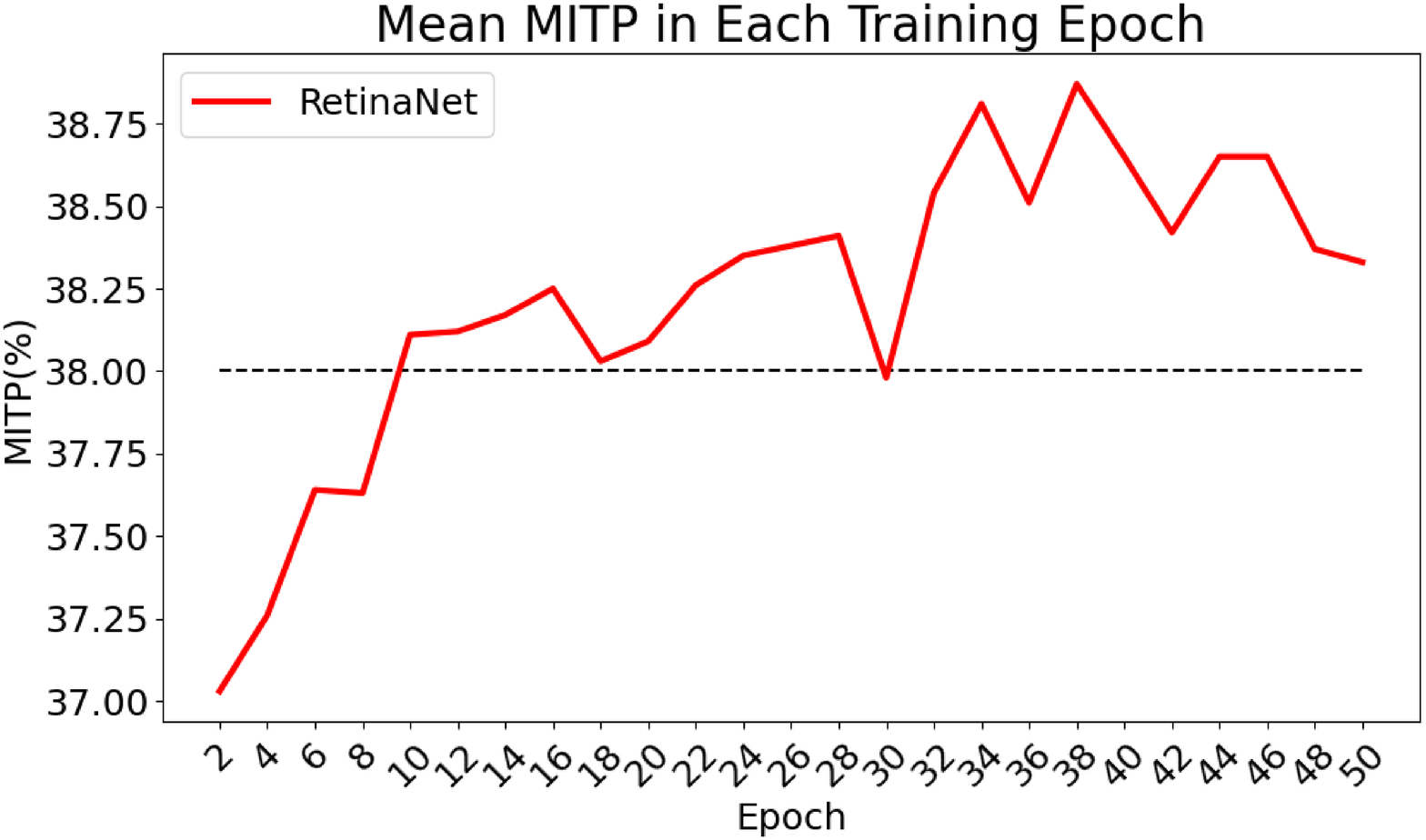} 
\caption{Mean MITP in training process}
\label{fig2}
\end{figure}

The reason that some anchors deviate from the preassigned objects is explained in Fig3. When the objects are sparse and do not overlap others, as shown in Fig3(a), the anchor will naturally regress to the unique object it could choose, as the green arrows indicated. However, when objects become crowded, something conflicting with the pattern in Fig3(a) will happen. As shown in Fig3(b), it is obvious that the IoU between the anchor and object B is greater than that between the anchor and object A, so in the way of general anchor assignment method, the anchor will be assigned to object B, and the optimizer will force the anchor to regress to object B, as the green arrows shown. But when A occurs in the position in Fig3(b), this forced optimization will conflict with the regression pattern learned from the situation in Fig3(a). With these conflicts existing, there will be three possible negative results during the training process: 1) if the regression direction is pulled back forcibly to object B, the prediction precision of the objects similar to the Fig3(a) might be impacted. 2) if the detector insists on regressing to object A, the anchor drift will happen. Object B might be ignored if this anchor is the only one assigned to it, which will decrease the Recall rate. 3) if the two forces in 1) and 2) cannot beat each other, the prediction will be ambiguous between object A and object B and not compact, which is the most common in actual. The compactness will influence the Missing Rate, which is sensitive to the false-positive rate. 

\begin{figure}[t]
\centering
\includegraphics[width=0.8\columnwidth]{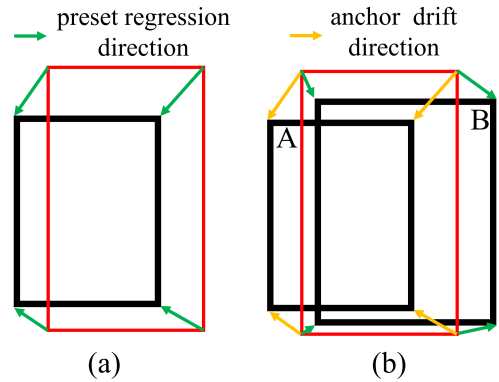} 
\caption{Causes of Anchor Drift Phenomenon}
\label{fig3}
\end{figure}

It is clear that the above three possible bad results are caused by the stiff original anchor-object assignment method, which directly assigns the anchors to the object with which they overlap the most despite the actual location and occlusion relationship of the crowded objects.

\section{Two-stage Anchor Assignment Method}

In this section, we propose a simple but effective anchor assignment method to assign the anchor boxes to objects adaptively.
\begin{figure*}[t]
\centering
\includegraphics[width=1.0\textwidth]{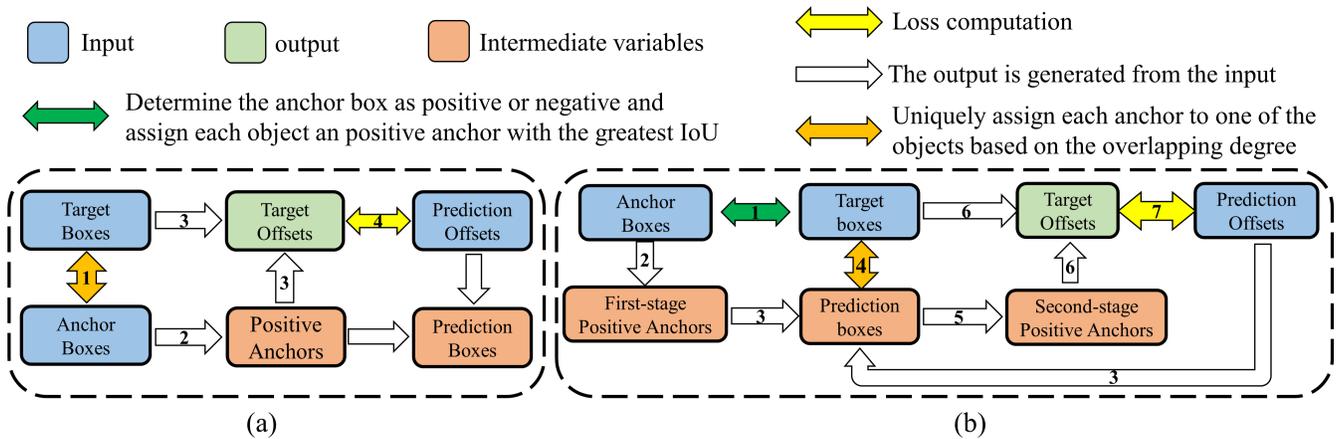} 
\caption{Flow chart of anchor-object assignment, the numbers on the arrows represent the flow sequence. (a) General anchor-object assignment method adapted by most SOTA detectors. (b) Two-stage anchor assignment method.}
\label{fig3}
\end{figure*}

According to the analysis above, the \textit{anchor drift} is led by the inflexible traditional anchor assignment method, as shown in Fig 4(a). The IoU values between the anchor boxes and object boxes are calculated firstly, and then these anchors are assigned to the objects with the largest IoU uniquely, as the dark yellow arrow labeled with 1 shown in Fig4(a). In the training process, the regression targets will be defined by these matching pairs, as the white arrow labeled with 3 shown in Fig4(a). These anchor boxes are preset manually. The hyper-parameters related to anchor boxes, such as scales and ratios, are always seen as shortcomings of the anchor-based detectors. They will directly determine which object each prediction corresponds to and will naturally influence the performance of detectors. 

In TSAA, the anchor assignment process is divided into two stages. In the first stage, the IoU values between all anchors and objects are calculated, and each anchor box is judged as positive or negative using a fixed IoU threshold as the generic anchor assignment method:

\begin{equation}
a=\left\{
\begin{array}{ll}
a_{p}& {\rm if}\ iou(a,t) \geq N_{t} \\
a_{n}& {\rm otherwise,} 
\end{array} \right.
\end{equation}
where $a$ is the anchor, $a_p$ is the positive anchor, $a_n$ is the negative anchor, $t$ is anyone object, and $N_t$ is the IoU threshold. 
To keep the Recall rate, each object is assigned one positive anchor with the largest IoU value. This process is shown as the green arrow labeled with 1 and the white arrow labeled with 2 in Fig4(b). We name the anchors that have been determined as positive as first-stage positive anchors (FSPA). As we know, each anchor box corresponds to a unique box prediction, so we restore all the prediction offsets whose corresponding anchors belong to the FSPA into the image coordinate and get the corresponding positive predicted boxes:
\begin{equation}
p_B=\left\{x_p,y_p,w_p,h_p\right\}=restore(a_p,p_{offset})
\end{equation}
in which $a_p=\left\{x_a,y_a,w_a,h_a\right\}$, $p_{offset}=\left\{t_x,t_y,t_w,t_h\right\}$ is the prediction offset corresponding to $a_p$, and the $x,y,w,h$ denote the box’s center point and its width and height.Different detectors have diverse restore functions since they adopt different training target building methods. For RetinaNet and Faster-RCNN, the restore functions $restor$e are:
\begin{equation}
\begin{array}{c}
x_{p}=t_{x} w_{a}+x_{a}, y_{p}=t_{y} h_{a}+y_{a} \\
w_{p}=e^{t_{w}} w_{a}, h_{p}=e^{t_{h}} h_{a}
\end{array}
\end{equation}
For YOLOv3, the restore functions are:
\begin{equation}
\begin{array}{c}
x_{p}=\sigma\left(t_{x}\right)+x_{a}, y_{p}=\sigma\left(t_{y}\right)+y_{a} \\
w_{p}=e^{t_{w}} w_{a}, h_{p}=e^{t_{h}} h_{a}
\end{array}
\end{equation}
In RetinaNet and Faster-RCNN, $(x_a,y_a)$ is the center point coordinate of the anchor box; in YOLOv3, $(x_a,y_a)$ is the coordinate of the top-left point of the cell grid where the object’s center point is located. The restore process is shown as the white arrows labeled with 3 in Fig4(b).

In the second stage, we assign these restored positive prediction boxes $p_B$ to one of the objects uniquely which have the largest IoU with these positive prediction boxes:
\begin{equation}
I=iou\left(P_{B}, T\right)
\end{equation}
where the $T=\left\{t_1,t_2,...,t_M\right\}$ is the ground-truth boxes, $P_B=\left\{p_{B1},p_{B2},...,p_{BN}\right\}$ is the set of positive prediction boxes, $iou(P_B,T)$ outputs a matrix with size $N\times M$ containing the IoU value of each $(p_{Bi},t_j)$ pair:
\begin{equation}
I_{ij}=iou\left(p_{Bi}, t_{j}\right)
\end{equation}
Along each row of $I$ we assign each prediction box to the object which overlaps it the most, as the dark yellow arrow labeled 4 shown in Fig4(b).Since the prediction boxes and anchors are in a one-to-one correspondence, we have also uniquely assigned each positive anchor box to one of the objects. We name these positive anchors which have been assigned to someone object the second-stage positive anchors(SSPA), as the white arrow labeled 5 shown in Fig3(b). Finally, with these object-anchor pairs, we obtain the training target offsets using the inverse operation of the restore function, as the white arrows labeled 6 shown in Fig4(b).

In TSAA, we assign each anchor by its corresponding prediction box, which means the network can participate in the assignment process rather than just depending on those manually preset and always-fixed anchors themselves to decide which objects to regress. Therefore, TSAA can avoid the ADP since it can ensure each anchor box directly match the object it will predict finally.

In the original anchor assignment method, to ensure the recall rate, each object will be assigned at least one anchor box with the largest IoU value, even if this largest IoU value is less than the IoU threshold. This mechanism is usually called low-quality-matching. In TSAA, we preserve this mechanism and adopt the original anchors rather than the prediction boxes to conduct the low-quality-matching step. This step is vital in TSAA. At the beginning of the training process, the network is initialized randomly, and the prediction boxes generated by the randomly initialized network are meaningless. If we only adopt the prediction boxes to decide the matching relationship, the network is hard to converge. Since the low-quality-matching relationship is fixed so that it can play the role of the teacher to guide the initial converge direction. In brief, the anchors selected by the low-quality-matching step are assigned to one of the targets fixedly in the whole training process, and the rest anchors are adaptively assigned according to their corresponding prediction boxes. 
\subsection{Variants toward Two-Stage Detectors}
The TSAA can both be integrated into the one-stage and two-stage detectors. However, there is something different between these two types of detectors when adopting the TSAA method. The two-stage detectors contain the RoI pooling, which crops the deep features to extract instance features. The proposal boxes generated by the RPN are crucial and dominate the instance feature extraction. Therefore, we use the proposal (anchor) boxes and their corresponding prediction boxes together to determine which objects each proposal (anchor) box should be assigned:
\begin{equation}
I_{ij}=iou\left(p_{Bi}, t_{j}\right)+iou\left(a_{i},t_{j}\right)
\end{equation}

In summary, the TSAA uses the prediction boxes in one-to-one correspondence with the anchor boxes to replace the manually preset anchor boxes in the original anchor assignment process. With the knowledge learned from training history integrated into each anchor box, each anchor box’s regression tendency will help them to find the most suitable object to regress. Hence, the conflict between the inflexible original anchor assignment method and the diversity of object distribution will be avoided naturally. The TSAA method is summarized in Algorithm 1. 

\begin{algorithm}[tb]
\caption{Two-stage anchor assignment method}
\label{alg:algorithm}
\textbf{Input}: $A$, $P$, $T$ \\
$A= \left \{a_1, a_2,...,a_L\right \}$ is the set of anchor boxes \\
$P= \left \{p_1, p_2,...,p_L\right \}$ is the set of prediction offsets \\
$T=\left \{t_1,t_2,...t_K\right \}$  is the set of objects \\
\textbf{Output}:$O$ is the set of target offsets corresponding positive anchor box \\
\textbf{Parameter}: $N_t$ is the IoU threshold \\

\begin{algorithmic}[1] 
\STATE Let $P_B=\left \{\right \}$, $A_B=\left \{\right \}$, $O=\left \{\right \}$,$M=\left\{\right\}$.
\FOR{$(a_j,p_j)$ in $(A,P)$}
\IF {$\mathop{\arg\max}\limits_{T} iou(a_j,T)\geq N_t$}
\STATE $P_B \leftarrow restore(a_j,p_j)$, $A_B \leftarrow a_j$
\ENDIF
\ENDFOR

\FOR {$t_i$ in $T$}
\STATE $a_m=\mathop{\arg\max}\limits_{A_B}iou(A_B,t_i)$
\STATE $O\leftarrow restore^{-1}(a_m,t_i)$ 
\STATE $M \leftarrow m$
\ENDFOR
\FOR {$p_{Bj}$ in $P_B$}
\IF {$j$ in $M$}
\STATE continue
\ELSE
\STATE $I^{at}_{\cdot j}=iou(a_j,T)$, $I^{pt}_{\cdot j}=iou(p_{Bj},T)$

\STATE $overlaps=I^{pt}_{\cdot j}$  (or $=I^{pt}_{\cdot j}+I^{at}_{\cdot j}$ for two-stage)

\STATE $t_{m}=\mathop{\arg\max}\limits_{T}overlaps$
\STATE $O\leftarrow restore^{-1}(a_j,t_{m})$
\ENDIF
\ENDFOR
\STATE \textbf{return} $O$
\end{algorithmic}
\end{algorithm}

\section{Experiment}

\subsection{Dataset and metrics}
An ideal anchor-object assignment method should be effective for crowded object detections and robust for generic (multi-class and less-crowded) object detection.

For crowded scenes, we adopt CrowdHuman to evaluate the effectiveness of TSAA. Compared with other datasets, CrowdHuman has much higher crowdedness. It contains 22.64 instances per image as well as 2.40 dense pair-wise (IoU$>$0.5) instances on average. There are three kinds of annotations in CrowdHuman, i.e., full-body, head, and visible. We only adopt full-body annotations.
For general scenes, we adopt COCO\cite{bib45}, which is the most convincing large-scale dataset in object detection. COCO contains 80 categories of instances and 9.34 instances per image on average. Using these two datasets, we can evaluate the TSAA in two kinds of extreme object distributions and demonstrate its robustness comprehensively.

We take the following three criteria as metrics to evaluate our method:

\textbf{AP(Average Precision)} is the most common metric in object detection, reflecting both the recall and precision ratios of the detection results. A larger AP means better performance.

\textbf{MR$^{-2}$} is the \textit{log-average Miss Rate on False Positive Per Image(FPPI)} in [$10^{-2}, 10^0$]. It’s sensitive to the false positive rate of detection results, and a smaller MR$^{-2}$ means better performance.

\textbf{JI(Jaccard Index)} evaluates how much the prediction box set overlaps the ground truth box set. In crowded object detection, JI is usually used to indicate the counting ability of a detector. A larger JI means better performance.

\subsection{Detectors and Detailed Settings}
We select three classic anchor-based detectors, whose anchor assignment methods are not exactly the same as each other, to cover as comprehensive anchor assignment methods as possible.

\textbf{RetinaNet} is a classic one-stage anchor-based detector. It judges an anchor as positive or negative by two fixed IoU thresholds, i.e., anchors boxes whose IoU value with any objects are greater than a fixed threshold $N_{t1}$ will be judged as positive samples; anchors boxes whose IoU value with any objects are less than another fixed threshold $N_{t2}$ will be judged as negative samples; the rest will be ignored. We set $N_{t1}$=$0.5$ and $N_{t2}$=$0.4$ in our experiment. We use ResNet50 pre-trained on ImageNet as the backbone of RetinaNet. The anchor scales are set the same as the \cite{bib20} with denser coverage situation, and set the aspect ratios $H$:$W$=$\left\{1:1, 1:2, 1:3\right\}$ for CrowdHuman and $\left\{2:1, 1:1, 1:2\right\}$ for COCO. For training, the batch size is set as 16, split into 4 GeForce RTX3090 GPUs. Each training runs for 50 epochs on CrowdHuman and 13 epochs for COCO. We optimize the detector using Stochastic Gradient Descent (SGD) with 0.0001 weight decay and 0.9 momenta. The initial learning rate is set to 0.005 for the first 32 epochs, 0.0005 for the next 10 epochs and 0.00005 for the last 8 epochs on CrowdHuman, and 0.01 for the first 8 epochs, 0.001 for next 3 epoch and 0.0001 for the last 2 epochs on COCO. The short edge of each training and testing image is resized to 800 pixels for both CrowdHuman and COCO. All testing results, including Faster-RCNN and YOLOv3 below, are generated with the NMS threshold of 0.5.

\textbf{Faster RCNN with FPN} is the most classic two-stage anchor-based detector. In the first stage, it adopts the same anchor assignment method as the RetinaNet. We set the $N_{t1}$=0.7 and $N_{t2}$=0.3 respectively. In the second stage, the proposals whose IoU values with any objects greater than a fixed threshold $N_{t3}$ will be judged as positive, we set $N_{t3}$=0.5, and the rest proposals are judged as negative. In experiments, we use TSAA only in the second stage to research the effectiveness of TSAA towards the anchors (proposals) generated by the RPN. Each training runs for 30 epochs on CrowdHuman and 11 epochs on COCO. The initial learning rate is set to 0.02 for the first 23 epochs, 0.002 for the next 4 epochs and 0.0002 for the last 3 epochs on CrowdHuman, and 0.02 for the first 8 epochs, 0.002 for the last 3 epochs on COCO. The anchor scales in RPN are the same as \cite{bib26}, and other settings are the same as the RetinaNet above. 

\textbf{YOLOv3} is another classic one-stage detector whose anchor assignment method is different from others. In original YOLOv3, an anchor box can be judged as positive only if two conditions are met simultaneously: 1) the center point of one of the object boxes is located in the feature cell grid that anchor corresponds; 2) the IoU between the anchor box and the object in condition 1) are greater than a fixed threshold.
\begin{figure}[t]
\centering
\includegraphics[width=0.5\columnwidth]{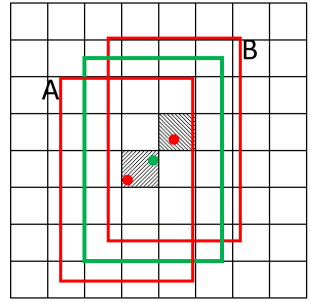} 
\caption{The difference in anchor assignment between RetinaNet and YOLOv3. The green frame is the target, and the red frame A and B are anchor boxes. In RetinaNet, frames A and B will all be assigned to the green target since their IoU values are greater than the threshold. In YOLOv3, however, only frame A will be assigned to the green target because the center point of frame B is out of the grid where the green box’s center point is located.}
\label{fig5}
\end{figure}
Fig5 shows the difference between RetinaNet and YOLOv3.
The anchor assignment method in YOLOv3 provides high-quality anchor boxes that are more aligned with objects spatially. In the improved YOLOv3, the condition 1) has another choice, i.e., the width ratio or height ratio between anchor and object is less than a fixed value $R_t$. To increase the variety of anchor assignment methods, we choose the improved condition 1) in the experiment and set $R_t$=4.0. The anchor judgment process is as follows:
\begin{equation}
a=\left\{\begin{array}{ll}
a_p& {\rm if} \max \left\{\frac{W_{a}}{W_{t}}, \frac{W_{t}}{W_{a}}, \frac{H_{a}}{H_{t}}, \frac{H_{t}}{H_{a}}\right\} \leq R_{t} \& \left\lfloor C_{a}\right\rfloor=\left\lfloor C_{t}\right\rfloor\\
a_n& {\rm otherwise,}
\end{array}\right.
\end{equation}
where the $W_a$ and $H_a$ are width and height of anchor, $W_t$ and $H_t$ are width and height of the target object, $C_a$ and $C_t$ are the center points’ coordinates of anchor and object, and $\left\lfloor x\right\rfloor$ gives the largest integer less than or equal to $x$. YOLOv3 uses the K-means algorithm to cluster the aspect ratios of objects in the dataset into 9 clusters. For training, the batch size is set to 96 on CrownHuman and 128 on COCO, split into 4 GPUs. On CrowdHuman, the image size is set to 640×640 pixels for training and 672×672 pixels for testing, and on COCO is set to $416\times 416$ for training and testing. Each training runs $T=300$ epochs. We optimize the detector using SGD with a momentum of 0.937, and a cosine learning rate scheduling strategy is adopted

\subsection{Ablation Studies on CrowdHuman}
In this section, ablation studies on CrowdHuman is conducted using the three classic detectors mentioned above. 

\subsubsection{RetinaNet}
\begin{table}[ht]
\centering
\begin{tabular}{l|l|l|l}
    Model & AP & MR$^{-2}$ & JI \\
    RetinaNet\cite{bib44} & 80.83 & 63.33 & $\slash$ \\
    RetinaNet(our impl) & 80.87 & 57.96 & 72.58 \\
    RetinaNet+TSAA & \textbf{81.18} & \textbf{53.51} & \textbf{74.03} \\
   
\end{tabular}
\caption{Ablation Studies on RetinaNet.}
\label{table1}
\end{table}

\begin{figure}[h]
\centering
\includegraphics[width=1.0\columnwidth]{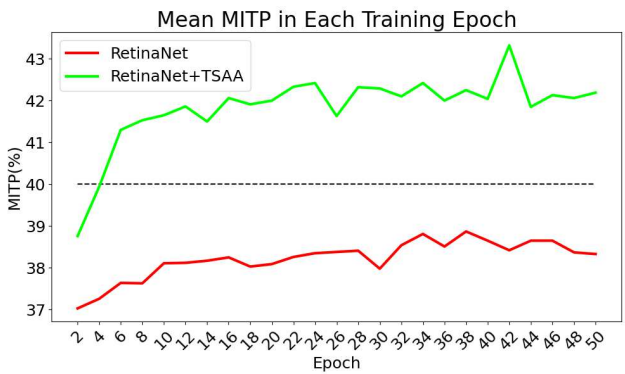} 
\caption{The MITP comparsion in RetinaNet with/without TSAA}
\label{fig6}
\end{figure}

\begin{figure*}[h]
\centering
\includegraphics[width=0.75\textwidth]{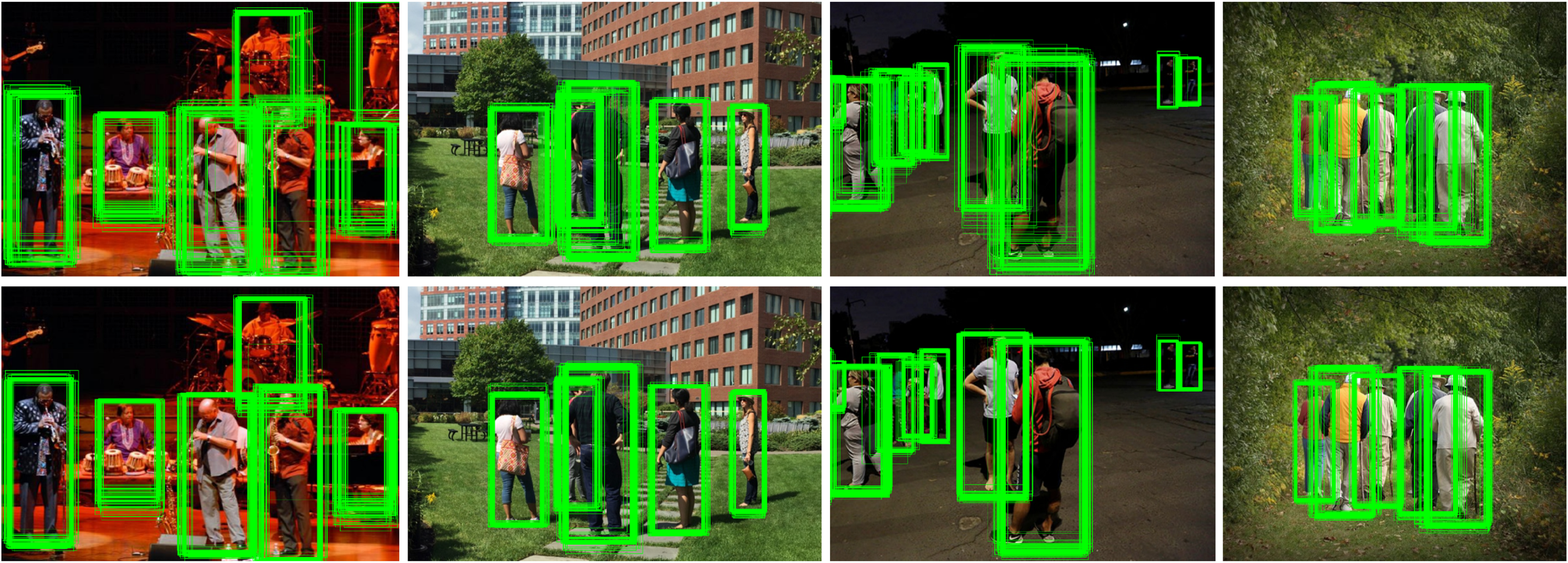} 
\caption{The visualization of detection results without the NMS process. The first row shows the baseline results, and the second row shows the results with TSAA.}
\label{fig7}
\end{figure*}

As shown in Table 1, with TASS, the MR$^{-2}$ is decreased by 4.45\%, JI is increased by 1.45\%, and AP is increased by 0.31\%. The MR$^{-2}$ is very sensitive to the false-positive rate, and the JI evaluates how much the prediction results overlap the ground truth, which means the ambiguous prediction boxes that hesitate between two objects will harm MR$^{-2}$ and JI. As analyzed, anchor drift in crowded scenes will make some prediction boxes ambiguous, so the improvements in MR$^{-2}$ and JI prove that TSAA can help to increase the compactness of the prediction boxes and suppress the false-positive prediction caused by ADP. The MR$^{-2}$ is also a critical metric in pedestrian detection. The significant improvements in MR$^{-2}$ brought by TSAA are very meaningful for pedestrian detection while increasing no extra computational cost and no extra network structures. In addition, AP is increased by 0.3\%, which means that TSAA also enhance the prediction precision.We also compare the mean MITP of each RetinaNet training epoch with/without TSAA, as shown in Fig6. With TSAA, the mean MITP in each epoch are significantly improved. The \textit{anchor drift} can seriously affect the IoU between the prediction box and its preassigned target box, i.e. let the prediction box drift to another target box, which will harm the MITP directly. In other words, the MITP is sensitive to the anchor drift, and the results shown in Fig6 indicte the \textit{anchor drift} can be suppressed by TSAA.

As shown in Fig7, the ambiguous boxes always appear in the middle area of two overlapped objects in baseline results. That is consistent with our analysis of the anchor drift phenomenon, i.e., the overlapped relationship between two objects sometimes makes the manually set optimization target contradicts the regression experience learned in the training process. With TSAA, it is evident that the compactness of prediction boxes, especially that of the overlapped objects, is significantly improved, and ambiguous prediction boxes are effectively suppressed.

\subsubsection{Faster-RCNN with FPN}
In this section, we explore the effectiveness of TSAA on the proposal boxes, which are not preset manually the same as RetinaNet but generated by RPN.  Table2 shows the experimental results.

\begin{table}[h]
\centering
\begin{tabular}{l|l|l|l}
    Model & AP & MR$^{-2}$ & JI \\
    F-RCNN\cite{bib44} & 83.1 & 52.4 & $\slash$ \\
    F-RCNN(our impl) & 87.17 & 43.76 & 79.38 \\
    F-RCNN+TSAA & \textbf{87.30} & \textbf{42.95} & \textbf{79.54} \\
   
\end{tabular}
\caption{Ablation Studies on Faster-RCNN with FPN.}
\label{table2}
\end{table}

From Table 2, TSAA reduces MR$^{-2}$ by 0.81\% and increases AP and JI by 0.13\% and 0.16\%, respectively, which suggests that the anchors generated by the RPN also have the possibility to drift to another surrounding object in crowded scenes. However, the improvements brought by TSAA in FPN are not as significant as in RetinaNet. We speculate the reason is the proposal boxes generated by RPN are more aligned to each object spatially, i.e., they are customized for each object by the RPN. Compared with the manually preset anchor boxes in one-stage detectors, the proposals are high-quality anchors for regression and the \textit{anchor drift} in the customized anchors(proposals) generated by RPN is not as common as the preset anchors in RetinaNet.

\subsubsection{YOLOv3}

Although the RetinaNet and YOLOv3 are all one-stage detectors, their anchor assignment methods are different. Compared with the RetinaNet, the anchor-target assignment method in YOLOv3 has an additional position constraint which makes the anchor and the object more center aligned spatially. Furthermore, the center point offset predictions in YOLOv3 are constrained within the feature cell grid, not as free as in RetinaNet, as equations (3) and (4) indicated.
The experimental results are shown in Table 3. 

\begin{table}[ht]
\centering
\begin{tabular}{l|l|l|l}
    Model & AP & MR$^{-2}$ & JI \\
    YOLOv3 & 85.14 & 50.27 & 72.08 \\
   YOLOv3+TSAA & \textbf{85.40} & \textbf{50.08} & \textbf{72.32} \\
   
\end{tabular}
\caption{Ablation Studies on CrowdHuman.}
\label{table3}
\end{table}

Table 3 shows that the TSAA method can comprehensively improve the performance of YOLOv3 in crowded object detection. Similar to the Faster-RCNN, the improvements in YOLOv3 are not as significant as that in RetinaNet. The AP and JI increased by 0.26\% and 0.24\%, respectively, and the MR$^{-2}$ decreased by 0.19\%. As mentioned above, the conditions in YOLOv3 to judge anchor as positive are stricter than in RetinaNet, which ensure its positive anchors are more center aligned to corresponding objects spatially. Intuitively, bad alignment in spatial-wise will exacerbate the \textit{anchor drift} and make detectors generate more ambiguous predictions. Similar to the Faster-RCNN, spatially alignment between anchors and objects guarantees the instance features and anchors are matched. The experimental results also prove that the original anchor assignment mechanism in YOLOv3 is better than that in RetinaNet. 

\subsection{Comparative Experiments}
To make the performance of TSAA comparable with the two-stage SOTA pedestrian detectors, we integrate TSAA into the Faster-RCNN-based pedestrian detector CrowdDet. The results of comparative experiments are listed in Table4. 

\begin{table}[h]
\centering
\begin{tabular}{l|l|l|l|l}
    Model  & Type & AP & MR$^{-2}$ & JI \\
    RFBNet\cite{bib37} & 1S & 78.33 & 65.22 & $-$ \\
    RetinaNet\cite{bib44} & 1S & 80.83 & 63.33 & $-$ \\
    AdaptiveNMS \cite{bib37}  & 1S & 79.67 & 63.03 & $-$ \\
    RetinaNet(our imple.) & 1S & 80.87 & 57.96 & 72.58 \\
    RetinaNet+TSAA &1S & 81.18 & 53.51 & \textbf{74.03} \\
    YOLOv3(our imple.) & 1S & 85.14 & 50.27 & 72.08 \\
    YOLOv3+TSAA & 1S & \textbf{85.40} & \textbf{50.08} & 72.32 \\
    FPN \cite{bib44} & 2S & 85.60 & 55.94 & $-$ \\
    AdaptiveNMS \cite{bib37} & 2S & 84.71 & 49.73 & $-$ \\
    FPN \cite{bib39} & 2S & 84.95 & 46.28 & $-$ \\
    Repulssion Loss \cite{bib41} & 2S & 85.64 & 45.69 & $-$ \\
    FPN \cite{bib38}  & 2S & 89.00 & 43.90 & $-$ \\
    PBN \cite{bib39} & 2S & 89.29 & 43.35 & $-$ \\
    FPN \cite{bib33} & 2S & 85.80 & 42.90 & 79.80 \\
    FPN(our imple.) & 2S & 87.17 & 43.76 & 79.38 \\
    FPN+TSAA & 2S & 87.30 & 42.95 & 79.54 \\
    CrowdDet\cite{bib33} & 2S & 90.70 & 41.40 & 82.30 \\
    CrowdDet+TSAA & 2S & \textbf{90.75} & \textbf{41.15} & \textbf{82.82} \\
   
\end{tabular}
\caption{Results of comparative experiment.1S means one-stage detector, 2S means two-stage detector.}
\label{table4}
\end{table}

From Table 4, we can see that, among all one-stage detectors, our YOLOv3-based detector achieves the best performance in AP and MR. And in all RetinaNet-based methods, our method also achieves the best performance, especially in MR, which goes a long way beyond others. Among all two-stage detectors, the CrowdDet with TSAA generates the best performance. The MR-2 is reduced by 0.25\%, and JI is increased by 0.52\%. In the set-NMS of CrowdDet, when the IoU between the proposal box and the candidate box is under the threshold (the threshold was set as 0.5 in CrowdDet), the set-NMS is equal to the traditional NMS, i.e., all boxes whose IoU with the current proposal box are under 0.5 will be preserved. That means if the boxes belonging to the same target are not compact enough, the false positive ratio will increase as well. With our TSAA, the compactness of the prediction box is increased, which suppresses the false positive brought by the loose boxes, and MR$^{-2}$ is reduced. On the other hand, the increased compactness means high precision. JI reflects how much the prediction boxes set overlaps the ground truth set. The JI is increased by 0.52\% means the prediction boxes are closer to the ground truth boxes, and the precision of the prediction box is improved. 

\subsection{Ablation Studies on COCO}
An advanced anchor assignment method should not only be customized for specific scenes but also be robust for generic circumstances. In this section, we conduct experiments on COCO.  A SOTA performance on COCO is not expacted, so we do not try to refine the detectors in terms of training skills or hyper-parameters to make them achieve SOTA-comparable performance. The purpose of ablation experiments on COCO is to study the impact of TSAA on generic object detection. Training methods and hyper-parameters are the same for the sake of fairness. Table5 shows the experimental results.

\begin{table}[htp]
\centering
\begin{tabular}{l|l|l|l}
    Model & AP & AP$_{50}$ & AP$_{75}$ \\
    RetinaNet(our impl) & 34.3 & 53.9 & 36.7 \\
    RetinaNet+TSAA & 34.7 & 54.2 & 37.3 \\
    F-RCNN(our impl) & 36.1 & 58.2 & 38.7 \\
    F-RCNN+TSAA & 36.2 & 58.0 & 39.3 \\
    YOLOv3(our impl) & 37.7 & 57.9 & 39.7 \\
    YOLOv3+TSAA & 37.9 & 58.0 & 39.9 \\
   
\end{tabular}
\caption{Ablation Studies on COCO}
\label{table4}
\end{table}

Without any changes in network structures and training methods, TSAA improves AP and AP$_{50}$ by 0.4\% and 0.3\% on RetinaNet, respectively.  Furthermore, the AP$_{75}$ is improved by 0.6\%, which is more distinct than AP and AP$_{50}$.
The improvements brought by TSAA in Faster-RCNN are not as significant as in RetinaNet, since more object-aligned proposals generated by the RPN can handle the sparse scenes well. However, we can still find that the AP$_{75}$ is increased by 0.6\%.  For YOLOv3, AP and AP$_{75}$ are improved by 0.2\%, and AP$_{50}$ is improved by 0.1\%. Among all results of these three detectors, we can find that the improvements in AP$_{75}$ are relatively more significant than other metrics, which means that the TSAA contributes to precision of the box regression. That is because the TSAA assists each proposal in finding the more suitable object to regress. In summary, the TSAA is robust to the generic object detection.

\begin{figure*}[p]
\centering
\includegraphics[width=1.0\textwidth]{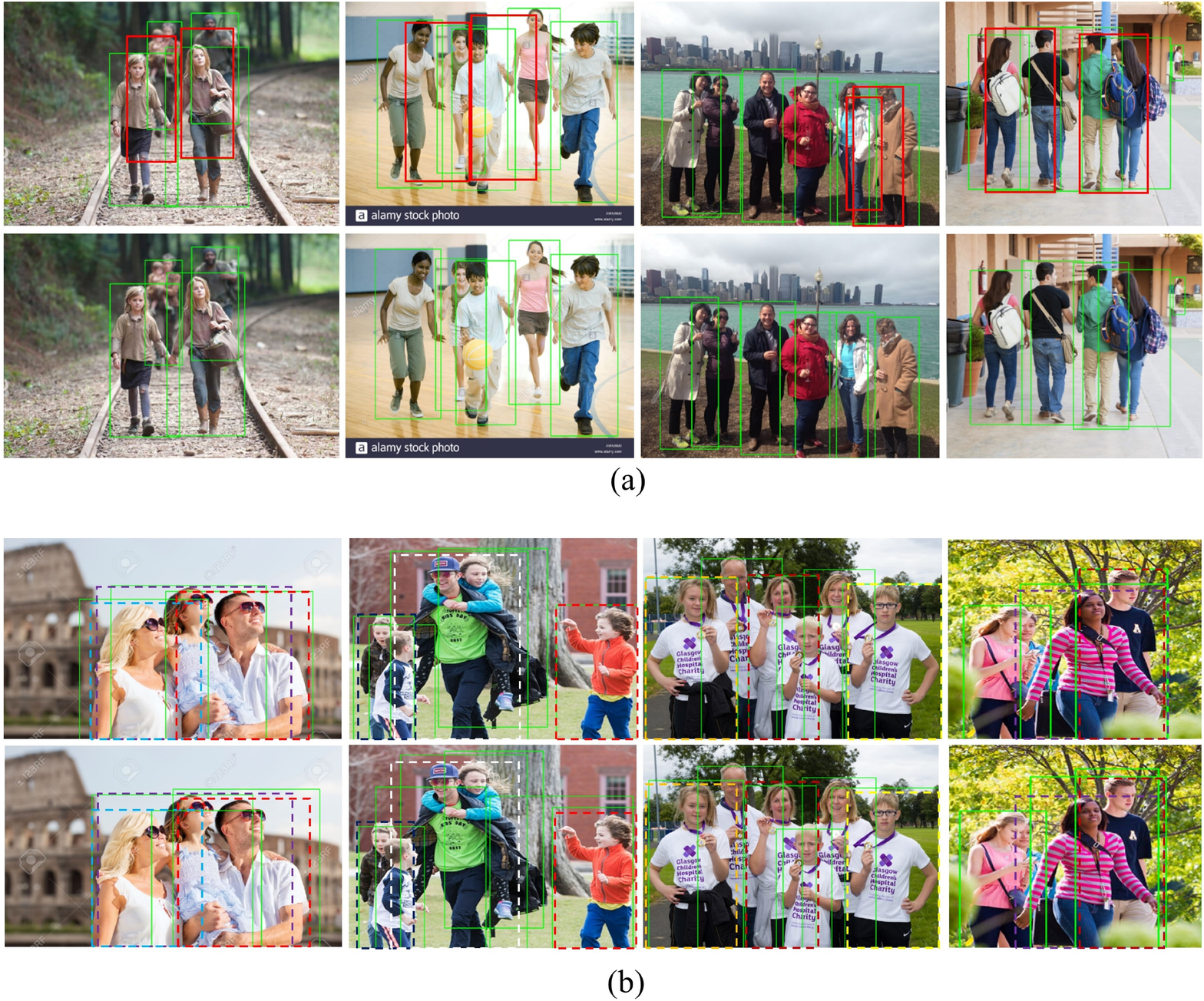} 
\caption{(a) Testing results of RetinaNet using NMS. The first row shows the baseline results. The second row shows the results with the TSAA method. To make it obvious, we re-print the false-positive boxes in red color.(b) Visualization of the improvement of the prediction precision in RetinaNet. The first row shows the results with TSAA, the second row shows the baseline results, in which the colorful dotted boxes are the ground truths, and the green boxes are the prediction results.}
\label{fig6}
\end{figure*}

\subsection{Qualitative Results}
We visualize the testing results of RetinaNet, as shown in Fig8. As shown by the red frames in Fig8(a), false-positive predictions are prone to occur between two objects in the original RetinaNet. This phenomenon is common in the whole testing set according to our observation. With the TSAA, the false-positive predictions can be effectively suppressed. Besides, the TSAA can also improve the precision of the prediction boxes. In the case where objects in the image are all detected, as the first row in Fig8(b) shown, the results with TSAA tend to be more precise compared with the baseline results in the second row in Fig8(b). This fact indicates that TSAA is quite usable in detection tasks that require high accuracy of box regression.

\section{Conculsion}

This paper defines and introduces the \textit{anchor drift} problem in anchor-based detectors in crowded scenes. In crowded scenes, the general anchor-object matching mechanism, which directly assigns each anchor to the object with the largest IoU value, usually generates an ambiguous prediction and leads to a high false-positive rate. To address this, an adaptive two-stage anchor assignment method TSAA is proposed to assign each anchor to the object which overlaps its corresponding prediction box the most. In this way, the network can participate in the anchor assignment process, i.e., adaptively assigning each anchor to the object the network thinks it should be assigned. Consequently, the optimization targets will be consistent with the actual regression targets, and the ambiguous prediction can be avoided. The experimental results show that TSAA can effectively suppress ambiguous predictions and reduce the miss rate in crowded scenes without affecting the performance of generic object detection.

\bibliography{aaai23}

\begin{thebibliography}{44}
\providecommand{\natexlab}[1]{#1}

\bibitem[{Bochkovskiy, Wang, and Liao(2020)}]{bib18}
Bochkovskiy, A.; Wang, C.; and Liao, H.~M. 2020.
\newblock YOLOv4: Optimal Speed and Accuracy of Object Detection.
\newblock \emph{CoRR}, abs/2004.10934.

\bibitem[{Bodla et~al.(2017)Bodla, Singh, Chellappa, and Davis}]{bib36}
Bodla, N.; Singh, B.; Chellappa, R.; and Davis, L.~S. 2017.
\newblock Soft-NMS — Improving Object Detection with One Line of Code.
\newblock In \emph{2017 IEEE International Conference on Computer Vision
  (ICCV)}, 5562--5570.

\bibitem[{Chen et~al.(2021)Chen, Li, Yuan, Su, and Li}]{bib9}
Chen, N.; Li, M.; Yuan, H.; Su, X.; and Li, Y. 2021.
\newblock Survey of pedestrian detection with occlusion.
\newblock \emph{Complex Intell. Syst.}, 7: 577--587.

\bibitem[{Chiang, Wang, and Chen(2021)}]{bib46}
Chiang, S.-H.; Wang, T.; and Chen, Y.-F. 2021.
\newblock Efficient pedestrian detection in top-view fisheye images using
  compositions of perspective view patches.
\newblock \emph{Image and Vision Computing}, 105: 104069.

\bibitem[{Chu et~al.(2020)Chu, Zheng, Zhang, and Sun}]{bib33}
Chu, X.; Zheng, A.; Zhang, X.; and Sun, J. 2020.
\newblock Detection in Crowded Scenes: One Proposal, Multiple Predictions.
\newblock In \emph{2020 IEEE/CVF Conference on Computer Vision and Pattern
  Recognition (CVPR)}, 12211--12220.

\bibitem[{Dai et~al.(2016)Dai, Li, He, and Sun}]{bib25}
Dai, J.; Li, Y.; He, K.; and Sun, J. 2016.
\newblock R-FCN: Object Detection via Region-based Fully Convolutional
  Networks.
\newblock In Lee, D.; Sugiyama, M.; Luxburg, U.; Guyon, I.; and Garnett, R.,
  eds., \emph{Advances in Neural Information Processing Systems}, volume~29.
  Curran Associates, Inc.

\bibitem[{Girshick(2015)}]{bib22}
Girshick, R. 2015.
\newblock Fast R-CNN.
\newblock In \emph{2015 IEEE International Conference on Computer Vision
  (ICCV)}, 1440--1448.

\bibitem[{Girshick et~al.(2014)Girshick, Donahue, Darrell, and Malik}]{bib21}
Girshick, R.; Donahue, J.; Darrell, T.; and Malik, J. 2014.
\newblock Rich Feature Hierarchies for Accurate Object Detection and Semantic
  Segmentation.
\newblock In \emph{2014 IEEE Conference on Computer Vision and Pattern
  Recognition}, 580--587.

\bibitem[{Hasan et~al.(2021)Hasan, Liao, Li, Akram, and Shao}]{bib10}
Hasan, I.; Liao, S.; Li, J.; Akram, S.~U.; and Shao, L. 2021.
\newblock Generalizable Pedestrian Detection: The Elephant In The Room.
\newblock In \emph{2021 IEEE/CVF Conference on Computer Vision and Pattern
  Recognition (CVPR)}, 11323--11332.

\bibitem[{He et~al.(2017)He, Gkioxari, Dollár, and Girshick}]{bib23}
He, K.; Gkioxari, G.; Dollár, P.; and Girshick, R. 2017.
\newblock Mask R-CNN.
\newblock In \emph{2017 IEEE International Conference on Computer Vision
  (ICCV)}, 2980--2988.

\bibitem[{Hei and Deng(2020)}]{bib30}
Hei, L.; and Deng, J. 2020.
\newblock CornerNet: Detecting Objects as Paired Keypoints.
\newblock \emph{International Journal of Computer Vision}, 128: 642--656.

\bibitem[{Huang et~al.(2020)Huang, Ge, Jie, and Yoshie}]{bib39}
Huang, X.; Ge, Z.; Jie, Z.; and Yoshie, O. 2020.
\newblock NMS by Representative Region: Towards Crowded Pedestrian Detection by
  Proposal Pairing.
\newblock In \emph{2020 IEEE/CVF Conference on Computer Vision and Pattern
  Recognition (CVPR)}, 10747--10756.

\bibitem[{hui Xu et~al.(2021)hui Xu, qing Wang, Wang, guo Duan, and Rui}]{bib2}
hui Xu, H.; qing Wang, X.; Wang, D.; guo Duan, B.; and Rui, T. 2021.
\newblock Object detection in crowded scenes via joint prediction.
\newblock \emph{Defence Technology}.

\bibitem[{Jiang et~al.(2018)Jiang, Luo, Mao, Xiao, and Jiang}]{bib24}
Jiang, B.; Luo, R.; Mao, J.; Xiao, T.; and Jiang, Y. 2018.
\newblock Acquisition of Localization Confidence for Accurate Object Detection.
\newblock In Ferrari, V.; Hebert, M.; Sminchisescu, C.; and Weiss, Y., eds.,
  \emph{Computer Vision -- ECCV 2018}, 816--832. Cham: Springer International
  Publishing.
\newblock ISBN 978-3-030-01264-9.

\bibitem[{Ju et~al.(2021)Ju, Luo, Liu, and Luo}]{bib6}
Ju, M.; Luo, J.; Liu, G.; and Luo, H. 2021.
\newblock ISTDet: An efficient end-to-end neural network for infrared small
  target detection.
\newblock \emph{Infrared Physics \& Technology}, 114: 103659.

\bibitem[{Lin et~al.(2017)Lin, Dollár, Girshick, He, Hariharan, and
  Belongie}]{bib26}
Lin, T.-Y.; Dollár, P.; Girshick, R.; He, K.; Hariharan, B.; and Belongie, S.
  2017.
\newblock Feature Pyramid Networks for Object Detection.
\newblock In \emph{2017 IEEE Conference on Computer Vision and Pattern
  Recognition (CVPR)}, 936--944.

\bibitem[{Lin et~al.(2020)Lin, Goyal, Girshick, He, and Dollár}]{bib20}
Lin, T.-Y.; Goyal, P.; Girshick, R.; He, K.; and Dollár, P. 2020.
\newblock Focal Loss for Dense Object Detection.
\newblock \emph{IEEE Transactions on Pattern Analysis and Machine
  Intelligence}, 42(2): 318--327.

\bibitem[{Lin et~al.(2014)Lin, Maire, Belongie, Hays, Perona, Ramanan,
  Doll{\'a}r, and Zitnick}]{bib45}
Lin, T.-Y.; Maire, M.; Belongie, S.; Hays, J.; Perona, P.; Ramanan, D.;
  Doll{\'a}r, P.; and Zitnick, C.~L. 2014.
\newblock Microsoft COCO: Common Objects in Context.
\newblock In Fleet, D.; Pajdla, T.; Schiele, B.; and Tuytelaars, T., eds.,
  \emph{Computer Vision -- ECCV 2014}, 740--755. Cham: Springer International
  Publishing.

\bibitem[{Liu, Han, and Rong(2021)}]{bib48}
Liu, G.; Han, J.; and Rong, W. 2021.
\newblock Feedback-driven loss function for small object detection.
\newblock \emph{Image and Vision Computing}, 111: 104197.

\bibitem[{Liu, Huang, and Wang(2019)}]{bib37}
Liu, S.; Huang, D.; and Wang, Y. 2019.
\newblock Adaptive NMS: Refining Pedestrian Detection in a Crowd.
\newblock In \emph{2019 IEEE/CVF Conference on Computer Vision and Pattern
  Recognition (CVPR)}, 6452--6461.

\bibitem[{Liu et~al.(2016)Liu, Anguelov, Erhan, Szegedy, Reed, Fu, and
  Berg}]{bib19}
Liu, W.; Anguelov, D.; Erhan, D.; Szegedy, C.; Reed, S.; Fu, C.-Y.; and Berg,
  A.~C. 2016.
\newblock SSD: Single Shot MultiBox Detector.
\newblock In \emph{Computer Vision -- ECCV 2016}, 21--37. Cham: Springer
  International Publishing.

\bibitem[{Ma et~al.(2022)Ma, Li, Zhu, Jiao, Tang, Guo, and Hou}]{bib5}
Ma, W.; Li, N.; Zhu, H.; Jiao, L.; Tang, X.; Guo, Y.; and Hou, B. 2022.
\newblock Feature Split–Merge–Enhancement Network for Remote Sensing Object
  Detection.
\newblock \emph{IEEE Transactions on Geoscience and Remote Sensing}, 60: 1--17.

\bibitem[{Ma et~al.(2020)Ma, Tian, Xu, Huang, and Li}]{bib29}
Ma, W.; Tian, T.; Xu, H.; Huang, Y.; and Li, Z. 2020.
\newblock AABO: Adaptive Anchor Box Optimization for Object Detection via
  Bayesian Sub-sampling.
\newblock In Vedaldi, A.; Bischof, H.; Brox, T.; and Frahm, J.-M., eds.,
  \emph{Computer Vision -- ECCV 2020}, 560--575. Cham: Springer International
  Publishing.
\newblock ISBN 978-3-030-58558-7.

\bibitem[{Moran et~al.(2021)Moran, Jiangning, Guangqi, and Haibo}]{bib7}
Moran, J.; Jiangning, L.; Guangqi, L.; and Haibo, L. 2021.
\newblock A real-time small target detection network.
\newblock \emph{Signal, Image and Video Processing}, 15: 1265--1273.

\bibitem[{Redmon et~al.(2016)Redmon, Divvala, Girshick, and Farhadi}]{bib14}
Redmon, J.; Divvala, S.; Girshick, R.; and Farhadi, A. 2016.
\newblock You Only Look Once: Unified, Real-Time Object Detection.
\newblock In \emph{Proceedings of the IEEE Conference on Computer Vision and
  Pattern Recognition (CVPR)}.

\bibitem[{Redmon and Farhadi(2017)}]{bib15}
Redmon, J.; and Farhadi, A. 2017.
\newblock YOLO9000: Better, Faster, Stronger.
\newblock In \emph{Proceedings of the IEEE Conference on Computer Vision and
  Pattern Recognition (CVPR)}.

\bibitem[{Redmon and Farhadi(2018)}]{bib16}
Redmon, J.; and Farhadi, A. 2018.
\newblock YOLOv3: An Incremental Improvement.
\newblock \emph{CoRR}, abs/1804.02767.

\bibitem[{Ren et~al.(2017)Ren, He, Girshick, and Sun}]{bib13}
Ren, S.; He, K.; Girshick, R.; and Sun, J. 2017.
\newblock Faster R-CNN: Towards Real-Time Object Detection with Region Proposal
  Networks.
\newblock \emph{IEEE Transactions on Pattern Analysis and Machine
  Intelligence}, 39(6): 1137--1149.

\bibitem[{Rukhovich et~al.(2021)Rukhovich, Sofiiuk, Galeev, Barinova, and
  Konushin}]{bib34}
Rukhovich, D.; Sofiiuk, K.; Galeev, D.; Barinova, O.; and Konushin, A. 2021.
\newblock IterDet: Iterative Scheme for Object Detection in Crowded
  Environments.
\newblock In \emph{Structural, Syntactic, and Statistical Pattern Recognition},
  344--354. Cham: Springer International Publishing.

\bibitem[{Shao et~al.(2018)Shao, Zhao, Li, Xiao, Yu, Zhang, and Sun}]{bib44}
Shao, S.; Zhao, Z.; Li, B.; Xiao, T.; Yu, G.; Zhang, X.; and Sun, J. 2018.
\newblock CrowdHuman: {A} Benchmark for Detecting Human in a Crowd.
\newblock \emph{CoRR}, abs/1805.00123.

\bibitem[{Tian et~al.(2019)Tian, Shen, Chen, and He}]{bib32}
Tian, Z.; Shen, C.; Chen, H.; and He, T. 2019.
\newblock FCOS: Fully Convolutional One-Stage Object Detection.
\newblock In \emph{2019 IEEE/CVF International Conference on Computer Vision
  (ICCV)}, 9626--9635.

\bibitem[{Tong and Wu(2022)}]{bib47}
Tong, K.; and Wu, Y. 2022.
\newblock Deep learning-based detection from the perspective of small or tiny
  objects: A survey.
\newblock \emph{Image and Vision Computing}, 123: 104471.

\bibitem[{Wan, Liu, and Chan(2021)}]{bib11}
Wan, J.; Liu, Z.; and Chan, A.~B. 2021.
\newblock A Generalized Loss Function for Crowd Counting and Localization.
\newblock In \emph{2021 IEEE/CVF Conference on Computer Vision and Pattern
  Recognition (CVPR)}, 1974--1983.

\bibitem[{Wang, Bochkovskiy, and Liao(2020)}]{bib17}
Wang, C.~Y.; Bochkovskiy, A.; and Liao, H. Y.~M. 2020.
\newblock Scaled-YOLOv4: Scaling Cross Stage Partial Network.
\newblock \emph{CoRR}, abs/2011.08036.

\bibitem[{Wang et~al.(2018)Wang, Xiao, Jiang, Shao, Sun, and Shen}]{bib41}
Wang, X.; Xiao, T.; Jiang, Y.; Shao, S.; Sun, J.; and Shen, C. 2018.
\newblock Repulsion Loss: Detecting Pedestrians in a Crowd.
\newblock In \emph{2018 IEEE/CVF Conference on Computer Vision and Pattern
  Recognition}, 7774--7783.

\bibitem[{Xiang et~al.(2022)Xiang, Miao, Liu, Haibo, and Moran}]{bib35}
Xiang, L.; Miao, H.; Liu, Y.; Haibo, L.; and Moran, J. 2022.
\newblock SPCS: a spatial pyramid convolutional shuffle module for YOLO to
  detect occluded object.
\newblock \emph{Complex \& Intelligent Systems}.

\bibitem[{Yang et~al.(2022)Yang, Tang, Cheung, Zhang, Liu, Ma, and Jiao}]{bib4}
Yang, Y.; Tang, X.; Cheung, Y.-M.; Zhang, X.; Liu, F.; Ma, J.; and Jiao, L.
  2022.
\newblock AR2Det: An Accurate and Real-Time Rotational One-Stage Ship Detector
  in Remote Sensing Images.
\newblock \emph{IEEE Transactions on Geoscience and Remote Sensing}, 60: 1--14.

\bibitem[{Zhang et~al.(2021{\natexlab{a}})Zhang, Wang, Dayoub, and
  Sünderhauf}]{bib1}
Zhang, H.; Wang, Y.; Dayoub, F.; and Sünderhauf, N. 2021{\natexlab{a}}.
\newblock VarifocalNet: An IoU-aware Dense Object Detector.
\newblock In \emph{2021 IEEE/CVF Conference on Computer Vision and Pattern
  Recognition (CVPR)}, 8510--8519.

\bibitem[{Zhang et~al.(2018)Zhang, Wen, Bian, Lei, and Li}]{bib40}
Zhang, S.; Wen, L.; Bian, X.; Lei, Z.; and Li, S.~Z. 2018.
\newblock Occlusion-Aware R-CNN: Detecting Pedestrians in a Crowd.
\newblock In Ferrari, V.; Hebert, M.; Sminchisescu, C.; and Weiss, Y., eds.,
  \emph{Computer Vision -- ECCV 2018}, 657--674. Cham: Springer International
  Publishing.

\bibitem[{Zhang, Yang, and Schiele(2018)}]{bib8}
Zhang, S.; Yang, J.; and Schiele, B. 2018.
\newblock Occluded Pedestrian Detection Through Guided Attention in CNNs.
\newblock In \emph{2018 IEEE/CVF Conference on Computer Vision and Pattern
  Recognition}, 6995--7003.

\bibitem[{Zhang et~al.(2021{\natexlab{b}})Zhang, He, Li, Li, See, and
  Lin}]{bib12}
Zhang, Y.; He, H.; Li, J.; Li, Y.; See, J.; and Lin, W. 2021{\natexlab{b}}.
\newblock Variational Pedestrian Detection.
\newblock In \emph{2021 IEEE/CVF Conference on Computer Vision and Pattern
  Recognition (CVPR)}, 11617--11626.

\bibitem[{Zheng et~al.(2022)Zheng, Zhang, Zhang, Qi, and Sun}]{bib3}
Zheng, A.; Zhang, Y.; Zhang, X.; Qi, X.; and Sun, J. 2022.
\newblock Progressive End-to-End Object Detection in Crowded Scenes.

\bibitem[{Zhou et~al.(2020)Zhou, Zhou, Peng, Du, Sun, Guo, and Huang}]{bib38}
Zhou, P.; Zhou, C.; Peng, P.; Du, J.; Sun, X.; Guo, X.; and Huang, F. 2020.
\newblock NOH-NMS: Improving Pedestrian Detection by Nearby Objects
  Hallucination.
\newblock In \emph{Proceedings of the 28th ACM International Conference on
  Multimedia}, MM '20, 1967–1975. New York, NY, USA: Association for
  Computing Machinery.
\newblock ISBN 9781450379885.

\bibitem[{Zhou, Wang, and Kr{\"{a}}henb{\"{u}}hl(2019)}]{bib31}
Zhou, X.; Wang, D.; and Kr{\"{a}}henb{\"{u}}hl, P. 2019.
\newblock Objects as Points.
\newblock \emph{CoRR}, abs/1904.07850.

\end{thebibliography}

\end{document}